%% file: paper.tex
\pdfoutput=1
\documentclass[10pt,twocolumn,letterpaper]{article}

\usepackage{iccv}
\usepackage{times}
\usepackage{epsfig}
\usepackage{graphicx}
\usepackage{subcaption}
\usepackage{amsmath}
\usepackage{amssymb}
\usepackage{color}
\usepackage{soul}
\usepackage[linesnumbered, ruled]{algorithm2e}
\usepackage{multirow}
\usepackage{booktabs}
\usepackage{amsfonts}
\usepackage{siunitx}
\usepackage{subfloat}
\usepackage{caption}
\usepackage{float}
\usepackage{multirow}
\usepackage{tabularx}
\usepackage{threeparttable}
\usepackage[outdir=./]{epstopdf}
\usepackage{footmisc}
\usepackage[T1]{fontenc}
\usepackage[utf8]{inputenc}
\usepackage{authblk}
\input{macros}

\usepackage[pagebackref=true,breaklinks=true,letterpaper=true,colorlinks,bookmarks=false]{hyperref}

\iccvfinalcopy 

\def\iccvPaperID{403} 

\newcommand*{\affaddr}[1]{#1} 
\newcommand*{\affmark}[1][*]{\textsuperscript{#1}}
\newcommand*{\email}[1]{{#1}}

\begin{document}

\title{Attributes2Classname: A discriminative model for attribute-based \\ unsupervised zero-shot learning}

\author{%
Berkan Demirel\affmark[1,3], Ramazan Gokberk Cinbis\affmark[2], Nazli Ikizler-Cinbis\affmark[3]\\
\affaddr{\affmark[1]HAVELSAN Inc.}, \affaddr{\affmark[2]Bilkent University}, \affaddr{\affmark[3]Hacettepe University}\\
\email{\tt\small bdemirel@havelsan.com.tr}, \email{\tt\small gcinbis@cs.bilkent.edu.tr}, \email{\tt\small nazli@cs.hacettepe.edu.tr}
}

\maketitle

\input{abstract}

\input{intro}

\input{relwork}

\input{method}

\input{experiments}

\input{conclusions}

{\small
\bibliographystyle{ieee}
\bibliography{bibabbr,paper}
}
\end{document}

%% file: abstract.tex
\begin{abstract} 
We propose a novel approach for unsupervised zero-shot learning (ZSL) of classes based on their names. Most existing unsupervised ZSL methods aim to learn a model for directly comparing image features and class names. However, this proves to be a difficult task due to dominance of non-visual semantics in underlying vector-space embeddings of class names. To address this issue, we discriminatively learn a word representation such that the similarities between class and combination of attribute names fall in line with the visual similarity. Contrary to the traditional zero-shot learning approaches that are built upon attribute presence, our approach bypasses the laborious attribute-class relation annotations for unseen classes. In addition, our proposed approach renders text-only training possible, hence, the training can be augmented without the need to collect additional image data. The experimental results show that our method yields state-of-the-art results for unsupervised ZSL in three benchmark datasets. 
\end{abstract}

%% file: intro.tex
\section{Introduction}

Zero-shot learning (ZSL) enables identification of classes that are not seen before by means of transferring knowledge from seen classes to unseen classes. This knowledge transfer is usually done via utilizing prior information from various auxiliary sources, such as attributes (\eg
\cite{lampert13pami,farhadi09cvpr,rohrbach2011evaluating, al2015transfer, xian2016latent, al2016recovering, akata2015evaluation}), class hierarchies (\eg \cite{rohrbach2011evaluating}), vector-space embeddings of class names (\eg \cite{xian2016latent, akata2015evaluation, al2016recovering}) and textual descriptions of classes (\eg \cite{lei2015predicting, elhoseiny2013write}). Among these, attributes stand out as an excellent source
of prior information: (i) thanks to their visual distinctiveness, it is possible to
build highly accurate visual recognition models of attributes; (ii) being
linguistically descriptive, attributes can naturally be used to encode classes
in terms of their visual appearances, functional affordances or other
human-understandable aspects.

\begin{figure}[t]
\begin{center}
   \includegraphics[width=\linewidth]{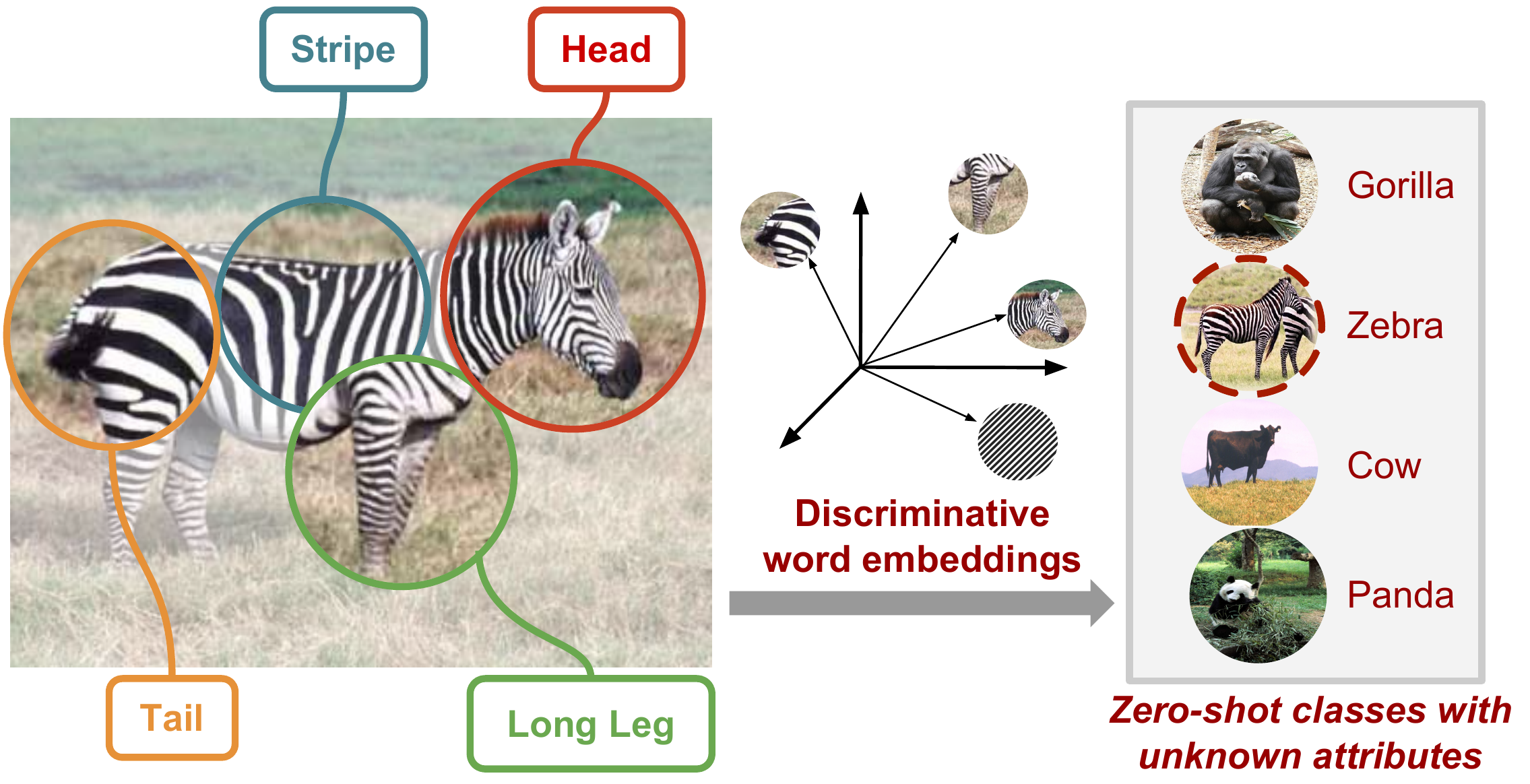}
\end{center}
   \vspace{-2mm}
\caption{We propose a zero-shot recognition model based on attribute and class names. Unlike most other attribute-based methods, our approach avoids the laborious attribute-class relations at test time, by discriminatively learning  
a word-embedding space for predicting the unseen class name, based on combinations of attribute names.}
\label{fig:framework}
\vspace{-2mm}
\end{figure}

Almost all attribute-based ZSL works, however, have an important disadvantage: attribute-class relations need to be
precisely annotated not only for the seen (training) classes, but also for the unseen (zero-shot) classes (\eg
\cite{farhadi09cvpr,lampert13pami,rohrbach2011evaluating,al2015transfer}).  This usually involves collecting
fine-grained information about attributes and classes, which is a time-consuming and error-prone task limiting the
scalability of the approaches to a great extent.

Several recent studies explore other sources of prior information to alleviate the need of collecting annotations at test time. These approaches rely on readily available sources like word
embeddings and/or semantic class hierarchies, hence, do not require dedicated annotation efforts. We simply refer to these as {\em unsupervised ZSL}. Such approaches, however, exclude attributes at the cost of exhibiting a lower recognition performance~\cite{akata2015evaluation}. 

Towards combining the practical merit of unsupervised ZSL with the recognition power of attribute-based methods, we
propose an attribute-based unsupervised ZSL approach. The main idea is to discriminatively learn a vector-space representation of words in which the combination of attributes relating to a class and the corresponding class name are mapped to nearby points. In this manner, the model would map distinctive attributes in images to a semantic word vector space, using which we can predict unseen classes solely based on their names. This idea is illustrated in Figure \ref{fig:framework}.

Our use of vector space word embeddings differs significantly from the way they are used in existing unsupervised ZSL
methods: existing approaches (\eg \cite{xian2016latent, akata2015evaluation}) aim to build a comparison function directly between image features and class
names. However, learning such a comparison function is difficult since word embeddings are likely to be dominated by
non-visual semantics, due to lack of visual descriptions in the large-scale text corpora that is used in the estimation of the embedding
vectors.
Therefore, the
resulting zero-shot models also tend to be dominated by non-visual cues, which can degrade the zero-shot recognition
accuracy. To address this issue, we propose to use the names of visual attributes as an intermediate layer that
connects the image features and the class names in an unsupervised way for the unseen classes.

An additional interesting aspect of our approach is the capability of \textit{text-only training}. Given pre-trained attribute models, the proposed ZSL model can be trained based on textual attribute-class associations, without the need for explicit image data even for training classes. This gives an extreme flexibility for scalability: the training set can be easily extended by enumerating class-attribute relationships, without the need for
collecting accompanying image data. The resulting ZSL model can then be used for recognition of zero-shot classes for which no prior attribute information or
visual training example is available.

We provide an extensive experimental evaluation on two ZSL object recognition and one ZSL action recognition benchmark
datasets. The results indicate that the proposed method yields state-of-the-art unsupervised zero-shot recognition
performance both for object and cross-domain action recognition. Our unsupervised ZSL model also provides competitive
performance compared to the state-of-the-art supervised ZSL methods. In addition, we experimentally demonstrate the
success of our approach in the case of text-only training. Finally, the qualitative results suggest that the non-linear
transformation of the proposed approach improves visual semantics of word embeddings, which can facilitate further
research. 

To sum up, our main contributions are as follows: (i) we propose a novel method for discriminatively learning a word vector space representation for relating class and attribute combinations purely based on their names. (ii) We show that the learned non-linear transformation improves the visual semantics of word vectors. (iii) Our method achieves the state-of-the-art performance among unsupervised ZSL approaches and (iv) we show that by augmenting the training dataset by additional class names and their attribute predicate matrices but no visual examples, a boost in performance can be achieved.

%% file: relwork.tex
\section{Related work}
\label{sec:relwork}

\vspace{2mm}

Initial attempts towards zero-shot classification were supervised, in the sense that they require explicit attribute annotations of the test classes (\eg \cite{lampert2009learning, lampert13pami, al2015transfer, rohrbach2011evaluating, deng2014large, jayaraman2014zero, romera2015embarrassingly, zhang2015zero, zhang2016zero, zhang2016zero1}). Lampert \textit{et al.}~\cite{lampert2009learning,lampert13pami} are among the first to use attributes in this setting. They propose direct (DAP) and indirect attribute prediction (IAP) where attribute and class relations are provided explicitly. Al-Halah \textit{et al.}~\cite{al2015transfer} introduce hierarchy and apply attribute label propagation on object classes, to utilize attributes at different abstraction levels. Rohrbach \textit{et al.}~\cite{rohrbach2011evaluating} propose a similar hierarchical method, but they use only class taxonomies. Deng \textit{et al.}~\cite{deng2014large} introduce Hierarchy and Exclusion (HEX) graphs as a standalone layer to be used on top of any-feedforward architecture for classification. 
Jayaraman and Grauman~\cite{jayaraman2014zero} propose a random forest approach to handle error tendencies of attributes. Romera \textit{et al.}~\cite{romera2015embarrassingly} develop two linear layered network to handle relations between classes, attributes and features. Zhang and Saligrama ~\cite{zhang2015zero} propose a method to use semantic similarity embedding where target classes are represented with histograms of the source classes.

An important limitation of the aforementioned methods is their dependency on the  attribute signatures of the test classes. To apply these approaches to additional unseen classes, the attribute signatures of those new classes need to be provided explicitly.  Our method alleviates this need by learning a word representation that allows zero-shot classification by comparing class names and attribute combinations,
with no explicit prior information about attribute relations
of unseen classes.

Recently, unsupervised ZSL methods are gaining more attention, due to their increased scalability. Instead of using class-attribute relations at test time, various auxiliary sources of side information, such as textual information~\cite{lei2015predicting, elhoseiny2013write} or word embeddings~\cite{akata2013label,akata2015evaluation,norouzi2013zero,frome2013devise, al2016recovering, changpinyo2016synthesized} are explored in such methods. Ba \textit{et al.}~\cite{lei2015predicting} propose to combine MLP and CNN networks handling text based information acquired from Wikipedia articles and visual information of images, respectively. Another interesting direction is explored by Elhoseiny \textit{et al.}~\cite{elhoseiny2013write}, where the classifiers are built directly on textual corpus that is accompanied with images.

Distributional word representations, or word embeddings, \cite{mikolov2013distributed,mikolov2013linguistic,pennington2014glove} are becoming increasingly popular \cite{akata2013label,akata2015evaluation,norouzi2013zero,frome2013devise}, due to the powerful vector-space representations where the distances can be meaningfully utilized. Akata \textit{et al.}~\cite{akata2013label} propose attribute label embedding (ALE) method that uses textual data as side information in the WSABIE~\cite{weston2010large} formulation. Akata \textit{et al.}~\cite{akata2015evaluation} improve ALE by using embedding vectors that were obtained from large-scale text corpora. Frome~\textit{et al.}~\cite{frome2013devise} propose a similar model where a pre-trained CNN model is fine-tuned in an end-to-end way to relate images with semantic class embeddings. Norouzi \textit{et al.}~\cite{norouzi2013zero} proposes to use convex combinations
of fixed class name embeddings, weighted by class posterior probabilities given by a pre-trained CNN model, to map images to the
class name embedding space. 
In the recent approach of Akata \textit{et al.}~\cite{akata2016multi} 
language representations are utilized jointly with the stronger supervision given by visual part annotations. Xian
\textit{et al.}~\cite{xian2016latent} use multiple visual embedding spaces to encode different visual characteristics of
object classes. Jain~\textit{et al.}~\cite{jain2015objects2action} and Kordumova~\textit{et al.}~\cite{kordumova2016pooling} leverage pre-trained object classifiers, and, action-object similarities given by class embeddings to assign action labels to unseen videos.



The work closest to ours is Al-Halah~\textit{et al.}~\cite{al2016recovering}, which proposes an approach for using   
visual attributes in the unsupervised ZSL setting. In their
approach, a model is learned to predict whether an individual attribute is related to a class name or not. For this purpose, they learn a separate bilinear compatibility function for each group of 
attributes, where similar attributes are grouped together to improve the performance. For unsupervised ZSL, 
this approach first estimates the association of attributes with the test
class, and then employs an attribute-based ZSL method using the estimated 
class-attribute relations. Our approach differs in two major ways.
First, instead of comparing classes with individual attribute names, we model the relationship between class names and combinations of attribute names. Second, as opposed to handling class-attribute relation estimation and zero-shot classification as two separate problems,  we discriminatively train our attribute based ZSL model in an end-to-end manner.

%% file: method.tex
\section{Method}
\label{sec:method}

In this section, we present the details of our approach. First, we explain our zero-shot learning model. Then, we describe how to train our ZSL model using discriminative {\em image-based training}
and {\em predicate-based training} formulations. Finally, we briefly discuss our {\em text-only training} strategy for
incorporating additional classes during training.

\input{method_zsl}

\input{method_ibt}

\input{method_pbt}

\subsection{Text-only training}

Predicate-based training, as explained in the previous section, is completely based on a class-attribute predicate
matrix for the training classes, and training images are used only for pre-training attribute classifiers that will be used at
test time. In contrast, image-based training, directly learns the ZSL model based on attribute classification
probabilities in training images, therefore in principle, we expect image-based training to perform better. This is,
in fact, verified in our experimental results: while predicate-based training shows competitive accuracy,
we obtain our state-of-the-art results using image-based training.

Despite the relatively lower performance of predicate-based training,
it has one interesting property: we can expand the training set by simply adding textual
information for additional novel classes into the predicate matrix. This allows improving the ZSL model by using classes with
no visual examples. We call incorporation of additional training classes in this manner as
{\em text-based training}. In Section~\ref{zero_exp}, we empirically show that it is possible to improve the
predicate-based training using text-based training.

%% file: method_zsl.tex
\subsection{Zero-shot learning model}
 
We define our ZSL model compatibility function $f(x,y): \mathcal{X}\times\mathcal{Y}\rightarrow\mathcal{R}$ that measures
the relevance of label $y \in \mathcal{Y}$ for a given image $x \in \mathcal{X}$.  Using this function, a test image $x$
can be classified simply by choosing the class maximizing the compatibility score: $\argmax_y f(x,y)$.

In order to enable zero-shot learning of classes based on class names only, we assume that an initial $d_0$-dimensional
vector space embedding $\varphi_y \in \mathcal{R}^{d_0}$ is available for each class $y$. These initial class name
embeddings are obtained using general purpose corpora, due to lack of a large-scale text corpus dedicated
for visual descriptions of objects.   The representations obtained by the class embeddings, hence, are typically
dominated by non-visual semantics. For instance, according to the GloVe vectors, the similarity between  {\em wolf} and
{\em bear} (both wild animals) is higher that the similarity between {\em wolf} and {\em dog}, though the latter pair is
visually much more similar to each other. 

These observations suggest that learning a compatibility function directly between the image features and class embeddings
may not be easy due to non-visual components of word embeddings.
To address this issue, we propose to leverage attributes, which are appealing for the dual representation they provide:
each attribute corresponds to (i) a visual cue in the image domain, and, (ii) a named entity in the language
domain, whose similarity with class names can be estimated using word embeddings. We define a function $\Phi(x): \mathcal{X} \rightarrow \mathcal{R}^d$ 
for embedding each image based on the attribute combination associated with it:
\begin{equation}
\Phi(x) = \frac{1}{\sum_a p(a|x)} \sum_a p(a|x) T( \varphi_a ) 
\label{eq:Phi}
\end{equation}
where $p(a|x)$ is the posterior probability of attribute $a$\footnote{The normalization in the denominator aims to make the embeddings comparable across images with varying number of observed attributes.}, given by a pre-trained binary attribute classifier,
$\varphi_a$ is the initial embedding vector of attribute $a$, and $T: \mathcal{R}^{d_0} \rightarrow \mathcal{R}^{d}$ is the transformation that we aim to learn.
Similarly, we define our class embedding function $\phi(y): \mathcal{Y} \rightarrow \mathcal{R}^d$ as the transformation
of the initial class name embeddings $\varphi_y$: $\phi(y) = T(\varphi_y)$.


The purpose of the function $T$ is to transform the initial word embeddings of attributes and classes
such that each image, and its corresponding class are represented by nearby points in the $d$-dimensional
vector embedding space. Consequently, we can define $f(x,y)$ as a similarity measure between the image and class embeddings.
In our approach, we opt for the cosine-similarity:
\begin{equation}
f(x,y) = \frac{\Phi(x)^\text{T} \phi(y)}{\|\Phi(x)\| \|\phi(y)\|} 
\label{eq:func_f}
\end{equation}
We emphasize that our approach requires only the name of an unseen class at test time,
as the compatibility function relies solely on the learned attribute and class name embeddings, rather than
attribute-class relations.

Figure~\ref{fig:method_test} illustrates our zero-shot classification approach. Given an image, we first apply the
attribute predictors and compute a weighted average of the attribute name embeddings. The class assignment
is done by comparing the resulting embedding of attribute combination with that of each (unseen) class name.
The image is then assigned to the class with the highest cosine similarity.

As defined above, the embeddings of attribute combinations and class names are functions of the shared
transformation $T(\varphi)$.\footnote{In principle, one can separately define a $T(\varphi)$ for attribute names, and, 
another one for class names. We have explored this empirically, but did not observe a consistent improvement.
Therefore, for the sake of simplicity, we use a shared transformation network in our experiments.}
In our experiments, we define $T(\varphi)$ as a two-layer feed-forward neural network.
In the following sections, we describe techniques for discriminatively learning this transformation network.

\input{fig_method_test}

%% file: fig_method_test.tex
\begin{figure}
\centering
\includegraphics[width=0.9\linewidth]{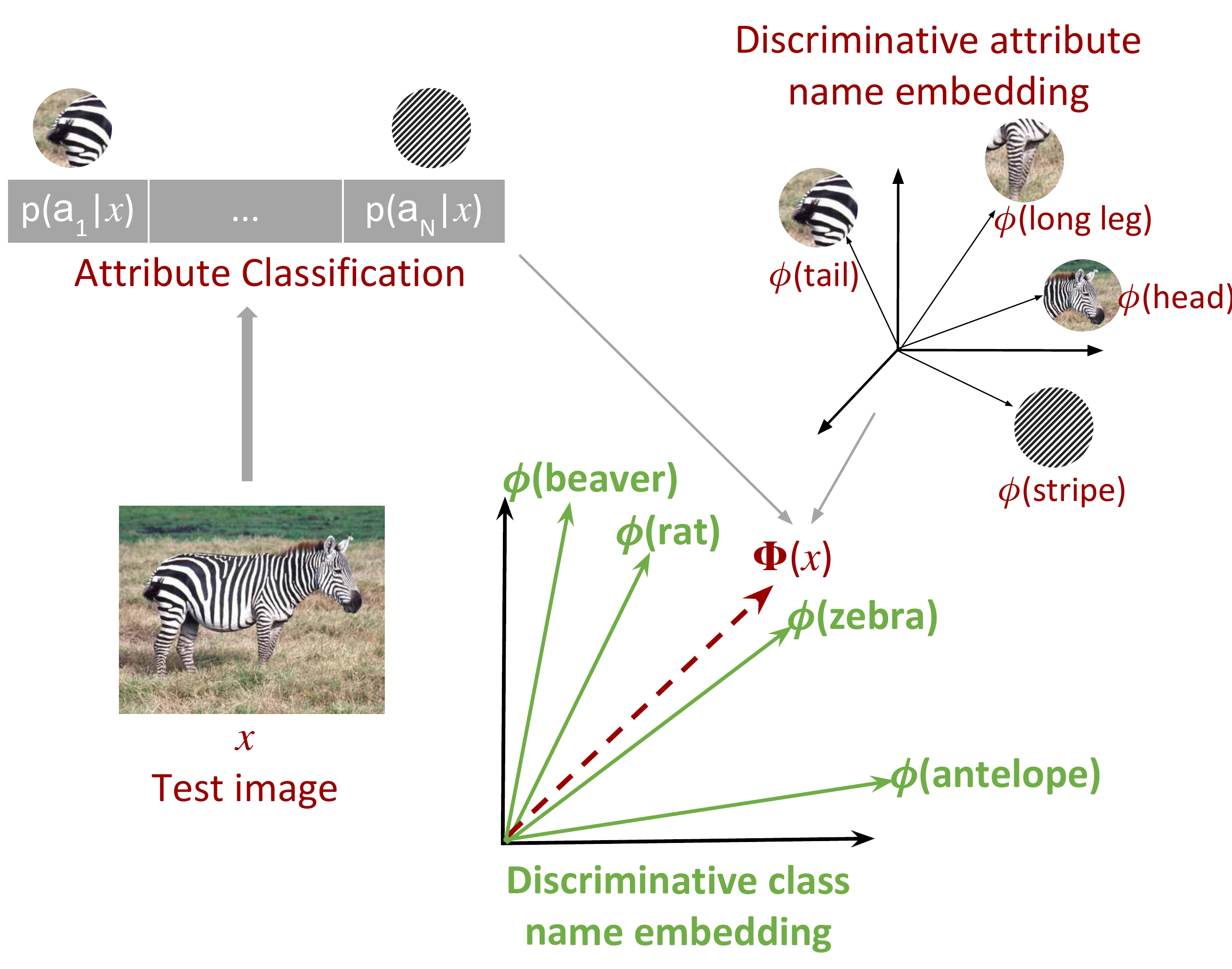}
\vspace{-1mm}
\caption{Illustration of our unsupervised zero-shot recognition model. Prediction depends on the similarity between  discriminatively learned representations of attribute combinations and class names. (Best viewed in color.)}
\label{fig:method_test}
\end{figure}

%% file: method_ibt.tex
\subsection{Image-based training (IBT)}

In image-based training, we assume that there exists a supervised training set $S$ of $N$ examples.
Each example forms an image and class label pair. By definition, no example in $S$ belongs to one of the zero-shot test classes. Our goal is to discriminatively learn the function $f(x,y)$ such that for each training example $i$,
the compatibility score of the correct class $y=y_i$ is higher than any other class $y_j$, by a margin of $\Delta(y_i,y_j)$.
More formally, the training constraint for the $i$-th training example is given by
\begin{align} 
    && f(x_i,y_i) \geq f(x_i,y_j)+\Delta(y_i,y_j), && \forall{y_j \neq y_i}
\label{eq:ibt_constraint}
\end{align} 
The margin function $\Delta$ indicates a non-negative pairwise discrepancy value for each pair of the training classes.

As explained in the previous section, $f(x,y)$ is a function of the transformation network $T(\varphi)$.
Let $\theta$ be the vector of all parameters in the transformation network.  Inspired from the structural
SVMs~\cite{tsochantaridis2005large,roller2004max}, we formalize our approach as a constrained optimization
problem:
\begin{equation}\begin{array}{cc}
    \min_{\theta,\xi} \lambda ||\theta|| + \sum_{i=1}^N \sum_{y_j  \neq  y_i} \xi_{ij} & \\
\vspace{-1mm} & \\
    f(x_i,y_i) \geq f(x_i,y_j) + \Delta(y_i,y_j) - \xi_{ij} & \forall y_j \neq y_i, \forall i
\end{array}\label{eq:ibt_constr_opt}\end{equation}
where $\xi$ is a vector of slack variables for soft-penalizing unsatisfied similarity constraints, 
and $\lambda$ is the regularization weight. To avoid optimization over non-linear constraints, we can equivalently
express this problem as an unconstrained optimization problem:
\begin{equation}\begin{array}{c}
\min_{\theta} \lambda \|\theta\|^2_2 + \\
\sum_{i=1}^N \sum_{y_j \neq y_i} \max\left(0,f(x_i,y_j) - f(x_i,y_i) +\Delta(y_i,y_j)\right)
\label{eq:ibt_unconst_opt}
\end{array}\end{equation}
Using this formulation, the transformation $T(\varphi)$ is learned in an discriminative and end-to-end manner, by
ensuring that the correct class score is higher than the incorrect ones, for each image.

We empirically observe that cross-validating the number of iterations provides an effective regularization
strategy, therefore, we fix $\lambda=0$. We use average Hamming distance between the attribute 
indicator vectors, which denote the list of attributes associated with each class, to compute $\Delta$ values.
This is the only point where we utilize the class-attribute predicate matrix in our image-based training approach.
In the absence of a predicate matrix, other types of $\Delta$ functions, like word embedding similarities, may be 
explored, which we leave for future work. Other implementation details are provided in Section~\ref{zero_exp}.

%% file: method_pbt.tex
\subsection{Predicate-based training (PBT)}

In this section, we propose an alternative training approach, which we call predicate-based training.
In this approach, the goal is to learn the ZSL model solely based on the predicate matrix, which
denotes the class-attribute relations. While image-based training is defined in terms of image-class similarities,
we formulate predicate-based training in terms of class-class similarities, without directly using any 
visual examples during training.

The predicate matrix consists of per-class indicator vectors,
where each element is one if the corresponding attribute is associated with the class, and zero, otherwise.
We denote the indicator vector for class $y$ by $\pi_y$. Then, similar to image embedding function $\Phi(x)$,
we define a {\em predicate-embedding} function $\Psi(\pi)$:
\begin{equation}
    \Psi(\pi) = \frac{1}{\sum_a \pi(a)} \sum_a \pi(a) T( \varphi_a ) .
\end{equation}
This embedding function is obtained by replacing posterior probabilities in \Eq{Phi} by binary
attribute-class relations. Then, we define a new compatibility function $g(\pi,y)$, 
as the cosine similarity between the vector $\Psi(\pi)$ and vector $\phi(y)$. This function is basically similar
to \Eq{func_f}, where the image embedding $\Phi(x)$ is replaced by the attribute indicator embedding $\Psi(\pi)$.

Finally, we define the learning problem as optimizing the function $g(x,y)$ such that for each class,
the compatibility score for its ideal set of attributes $\pi_y$ is higher than the attribute combination $\pi_{y^\prime}$
of another class $y^\prime$, by a margin of $\Delta(y,y^\prime)$. This constraint aims to ensure that
the similarity between the name embedding of a set of attributes and the embedding of a class name reliably indicates
the visual similarity indicated by the predicate matrix.

This definition leads us to an unconstrained optimization problem analogous to \Eq{ibt_unconst_opt}:
\begin{equation}\begin{array}{c}
\min_{\theta} \lambda \|\theta\|^2_2 + \\
\sum_{y=1}^K \sum_{y^\prime \neq y_i} \max\left(0,g(\pi_{y^\prime},y_i) - g(\pi_{y_i},y_i) +\Delta(y_i,y^\prime)\right)
\label{eq:pbt_unconst_opt}
\end{array}\end{equation}
where $K$ indicates the number of training classes in the predicate matrix. As in image-based training, we define
$\Delta(y,y^\prime)$ as the average Hamming distance between $\pi_y$ and $\pi_{y^\prime}$, and use $\lambda=0$.

Figure~\ref{fig:method_pbt} illustrates the predicate-based training approach. As shown in this figure, the main idea is
to project the $\varphi$ word representations into a new space, where the similarity between a class and an attribute
combination in terms of their name vectors is indicative of their visual similarity. At test time, we use 
the learned transformation network in zero-shot classification via the compatibility function $f(x,y)$ in \Eq{func_f}. This compatibility function uses only attribute classifier outputs and the transformed word embeddings.

\input{fig_method_pbt}

%% file: fig_method_pbt.tex
\begin{figure}
\centering
\includegraphics[width=\linewidth]{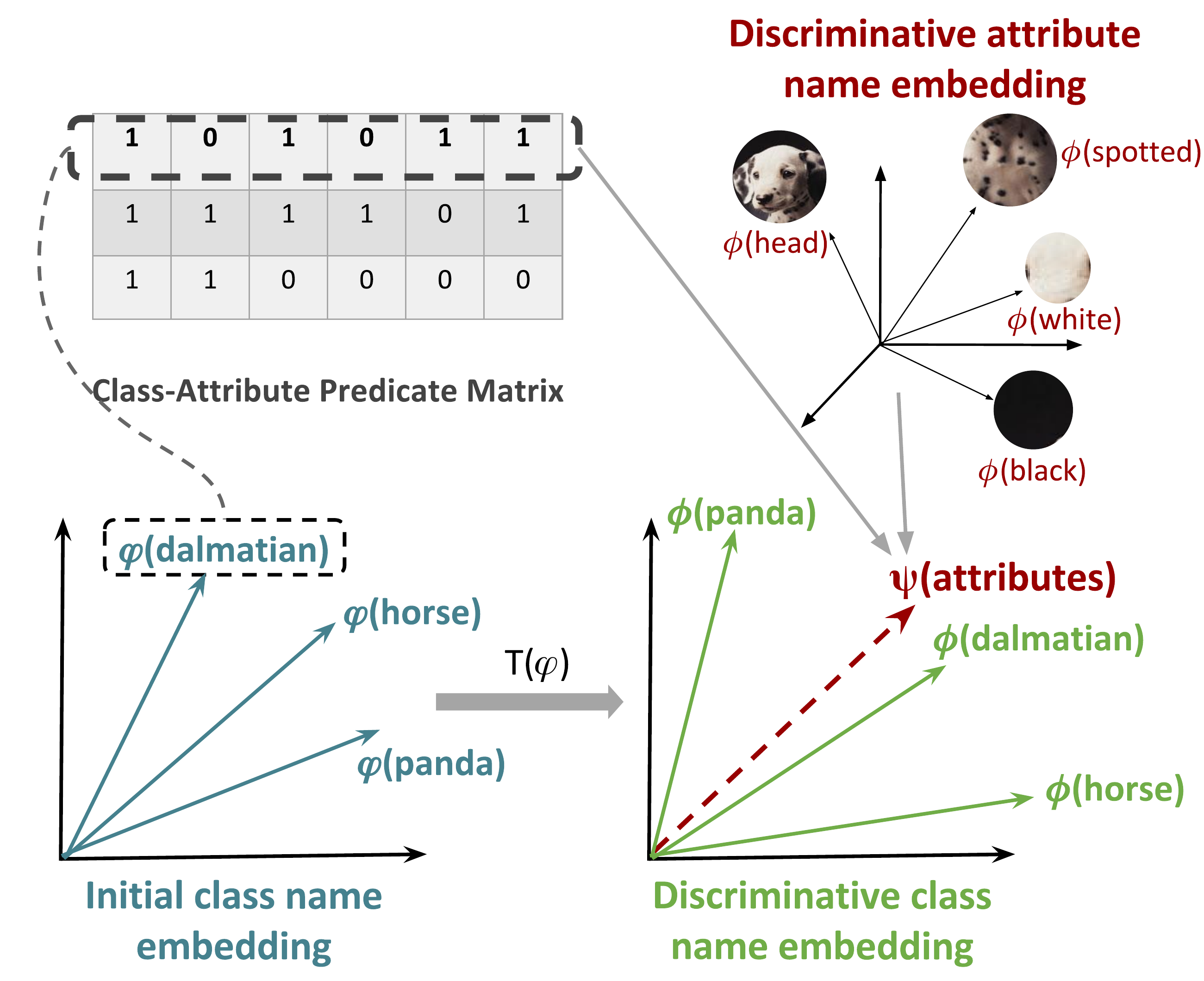}
\vspace{-4mm}
\caption{Illustration of our predicate-based training approach, which uses only the predicate matrix of class and attribute relations as the source of supervision. The goal is to represent class and attribute combinations, based on their names, in a space where each class is closest to its ideal attribute combination.}
\label{fig:method_pbt}
\end{figure}

%% file: experiments.tex
\section{Experiments} \label{zero_exp}
To evaluate the effectiveness of the proposed approach, we consider two different ZSL applications: zero-shot object classification and zero-shot action recognition.

\subsection{Zero Shot Object Classification}
In this part, we explain our zero-shot object classification experiments on two common datasets namely AwA~\cite{lampert13pami}, aPaY~\cite{farhadi2009describing}. AwA dataset~\cite{lampert13pami} contains 30,475 images of 50 different animal classes. 85 per-class attribute labels are provided in the dataset. In the predefined split for zero-shot learning, 40 animal classes are marked for training and 10 classes for testing. aPaY dataset~\cite{farhadi2009describing} is formed of images obtained from two different sources. aPascal (aP) part of this dataset is obtained from PASCAL VOC 2008~\cite{pascal-voc-2008}. This part contains 12,695 images of 20 different classes. The second part, aYahoo (aY), is collected using Yahoo search engine and contains 2,644 images of 12 object classes completely different from aPascal classes. Images are annotated with 64 binary per-image attribute labels. In zero-shot learning settings on this dataset, aPascal part is used for training and aYahoo part is used for testing. We follow the same experimental setup as in \cite{al2015transfer} and only use training split of aPascal part to learn attribute classifiers.

\vspace{2mm}
\noindent\textbf{Attribute Classifiers.} We use CNN-M2K features \cite{al2015transfer} to encode images and train attribute classifiers. We resize each image to 256x256 and then subtract the mean image. Data augmentation is applied via using five different crops and their flipped versions. Outputs of fc7 layer are used, resulting in 2,048 dimensional feature vectors. Following~\cite{farhadi2009describing}, we obtain the attribute classifiers by training $\ell_2$-regularized squared-hinge-loss linear SVMs. Parameter selection is done using 10-fold cross validation over the training set and Platt scaling is applied to map the attribute prediction scores to posterior probabilities. For image-based training, cross-validation outputs are used as the classification scores in training images.

\vspace{2mm}

\noindent\textbf{Word Embeddings.} For each class and attribute name, we generate a 300-dimensional word embedding vector using GloVe \cite{pennington2014glove} based on Common Crawl Data\footnote{\href{http://commoncrawl.org/the-data/}{~commoncrawl.org/the-data/}}. These word vectors are publicly available\footnote{\href{http://nlp.stanford.edu/projects/glove/}{~nlp.stanford.edu/projects/glove/}}. For those names that consist of multiple words, we use the average of the word vectors.

\vspace{2mm}
\noindent\textbf{Word Representation Learning.} We define the transformation function as a two layer feed-forward network. We use 2-fold cross-validation over the training set to select number of hidden units and number of iterations. 
\textit{tanh} function is used as the activation function in the first hidden layer and \textit{sigmoid} function is used in the second hidden layer. 
Adam \cite{kingma2014adam} is used for stochastic optimization, and learning rate value is set to 1e-4. Implementation is done using TensorFlow~\cite{abadi2015tensorflow}.\footnote{\href{https://github.com/berkandemirel/attributes2classname}{~github.com/berkandemirel/attributes2classname}}

\begin{table}
\begin{center}
\caption{Zero-shot classification performance of proposed predicate-based (PBT) and image-based (IBT) methods on AwA and aPaY datasets. We report normalized accuracy.} 
\label{table:ourMethods} 
\begin{tabular}{ c c c }
\hline
    Method & AwA & aPaY \\
\hline
    Baseline & 10.2 & 16.0\\
    PBT & 60.7 & 29.4\\
    IBT & \textbf{69.9} & \textbf{38.2} \\
\hline
\end{tabular}
\end{center}
\end{table}

\vspace{2mm}
\noindent\textbf{Results.} In our experiments, we first evaluate the performance of attribute classifiers, since this is likely to have a significant influence on zero-shot classification. The attribute classifiers yield 80.56\% mean AUC on the AwA dataset, 84.91\% mean AUC on the aPaY dataset. These results suggest that our attribute classifiers are relatively accurate, if not perfect. Further improvements in attribute classification are likely to have a positive impact on the final ZSL performance.

\begin{figure}
\centering
\includegraphics[trim={4cm 0 4cm 0},clip,width=0.45\textwidth ]{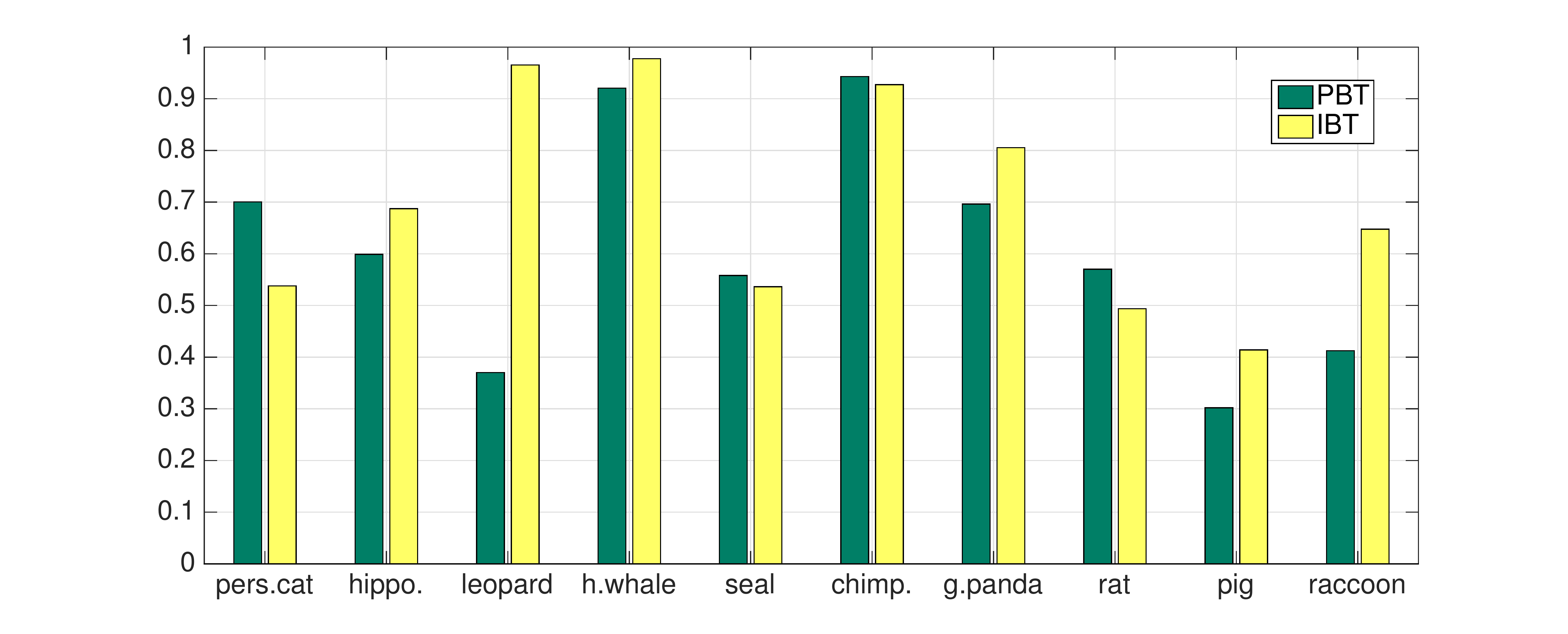}
\caption{Class-wise prediction accuracies on AwA Dataset. }
\label{fig:roc_awa}
\end{figure}

Table \ref{table:ourMethods} presents the experimental results for our approach. 
In this table, {\em baseline} represents the case where the transformation $T(\varphi)$ is defined as an identity mapping. PBT (predicate-based training) represents our proposed approach that learns a transformation using the attribute predicate matrix, whereas IBT (image-based training) represents learning transformation using training images. The results in Table \ref{table:ourMethods} shows the importance and success of our learning formulations, compared to the baseline. In addition, we observe that image-based training outperforms predicate-based training on average, which is in accordance with our expectations. Class-wise accuracy comparison of PBT and IBT methods is given in Figure~\ref{fig:roc_awa}. We observe that some of the classes respond particularly well to the image-based training.

Table \ref{table:comparisonSOA} presents a comparison of our results against
 a number of supervised and unsupervised ZSL methods. In this table, the supervision corresponds to the information needed during test time for zero-shot learning: the supervised methods require additional data about the unseen classes such as attribute-class predicate matrices, whereas unsupervised methods do not require any explicit inputs about test classes. Hence, supervised methods have a very major advantage in this comparison, as they employ external attribute signatures of test classes. In contrast, unsupervised methods carry out zero-shot classification among the test classes without using data additional to the training set. Finally, we note that, we exclude ZSL methods that operate on low-level visual image features, as their results are not directly comparable.  Instead, for the sake of fair comparison, we only compare to those methods that use similar convolutional neural network based image representations.

\begin{table}
\begin{center}
\caption{Comparison to state-of-the-art ZSL methods (unsupervised and supervised).} 
\label{table:comparisonSOA}
\begin{tabularx}{\linewidth}{ c l l l l l }
\hline
Test supervision & Method & AwA & aPaY \\
\hline
\multirow{8}{6em}{unsupervised} 
    &DeViSE\cite{frome2013devise}&44.5 & 25.5 \\
    &ConSE\cite{norouzi2013zero}&46.1 & 22.0 \\
    &Text2Visual\cite{elhoseiny2013write,bo2010twin}&55.3 & 30.2 \\
    &SynC\cite{changpinyo2016synthesized}&57.5&-\\
    &ALE\cite{akata2015evaluation} &58.8 & 33.3 \\
    &LatEm\cite{xian2016latent}&62.9 & - \\
    &CAAP\cite{al2016recovering}&67.5 & 37.0 \\
    &Our method & \textbf{69.9}& \textbf{38.2} \\
    \midrule
\multirow{9}{6em}{supervised} & 
    DAP\cite{lampert13pami} & 54.0& 28.5 \\
    & ENS\cite{rohrbach2011evaluating} & 57.4& 31.7 \\
    & HAT\cite{al2015transfer} & 63.1& 38.3 \\
    & ALE-attr\cite{akata2015evaluation} & 66.7& - \\
    & SSE-INT\cite{zhang2015zero} & 71.5& 44.2 \\
    & SSE-ReLU\cite{zhang2015zero} & 76.3& 46.2 \\
    & SynC-attr\cite{changpinyo2016synthesized}&76.3&-\\
    & SDL\cite{zhang2016zero} & 79.1& 50.4 \\
    & JFA\cite{zhang2016learning} & 81.0& 52.0 \\
\hline
\end{tabularx}
\end{center}
\vspace{-4mm}
\end{table}

From Table \ref{table:comparisonSOA} we see that on AwA and aPaY datasets, our unsupervised ZSL method yields state-of-the-art classification performance compared to other unsupervised ZSL methods. In addition, our method performs on par with some of the supervised ZSL methods.

\subsection{Zero Shot Action Recognition}

For zero-shot action recognition, we evaluate our approach on UCF-Sports Action Recognition Dataset~\cite{soomro2014action}. The dataset is formed of videos from various sport actions which are featured from television channels such as the BBC and ESPN, and contains a total of 150 videos of 10 different sport action classes. 

\vspace{2mm}
\noindent\textbf{Word Embeddings.} Following \cite{jain2015objects2action}, we utilize 500-dimensional word embedding vectors generated with the skip-gram model of word2vec~\cite{mikolov2013distributed} learned over YFCC100M~\cite{thomee2015new} dataset. YFCC100M dataset contains metadata tags of about 100M Flickr images and the word vectors obtained from YFCC100M are publicly available\footnote{\label{note1}\href{https://staff.fnwi.uva.nl/m.jain/projects/Objects2action.html}{~staff.fnwi.uva.nl/m.jain/projects/Objects2action.html}}. 

\vspace{2mm}
\noindent\textbf{Object Classifiers.} Since there is no explicit definition of attributes for actions, the object cues can be leveraged instead of attributes, as suggested by \cite{jain2015objects2action}. To this end, we obtain predicate matrices from the textual data by measuring the cosine similarity between actions and object classification scores. We operate on the object classification responses made available by \cite{jain2015objects2action}\footref{note1}. These are obtained by AlexNet\cite{krizhevsky2012imagenet}, where every 10th frame is sampled for each video and each sampled frame is represented with the total of 15,293 ImageNet object categories. Average pooling is applied afterwards, so that each video is represented with 15,293 dimensional vectors. To have a fair comparison, we also apply the sparsification step of \cite{jain2015objects2action} using the same parameters. This sparsification is done for eliminating noisy object classification responses.

\vspace{2mm}
\noindent\textbf{Word Representation Learning.} Model learning settings are the same with those of ZSL object classification experiments, with the exception that only image-based loss is used, because predicate matrices are not available during training. Since we do not have any training data for target datasets, we train our transformation function with a different dataset (\textit{i.e.} UCF-101~\cite{ucf101}). To avoid any overlap between datasets, we exclude the common action classes from the training set for an accurate zero-shot setting. Some of such common classes that are excluded from training are  \textit{Diving} and \textit{Horse Riding}.
\newcommand{\rulesep}{\unskip\ \vrule\ }

\begin{figure*}
\captionsetup[subfigure]{labelformat=empty}
\centering

\begin{subfigure}[b]{1.4cm}
\includegraphics[width=\linewidth,height=1.4cm]{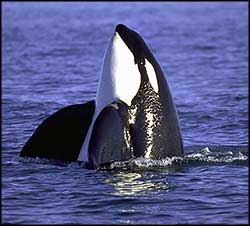}
\caption{K. Whale}
\end{subfigure}
\hspace{.3cm}
\rulesep
\hspace{.3cm}
\begin{subfigure}[b]{1.4cm}
\includegraphics[width=\linewidth,height=1.4cm]{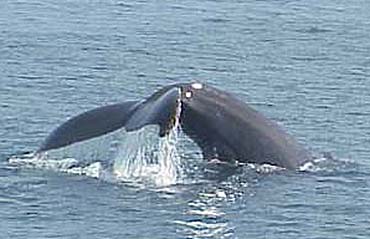}
\caption{B. Whale}
\end{subfigure}
\begin{subfigure}[b]{1.4cm}
\includegraphics[width=\linewidth,height=1.4cm]{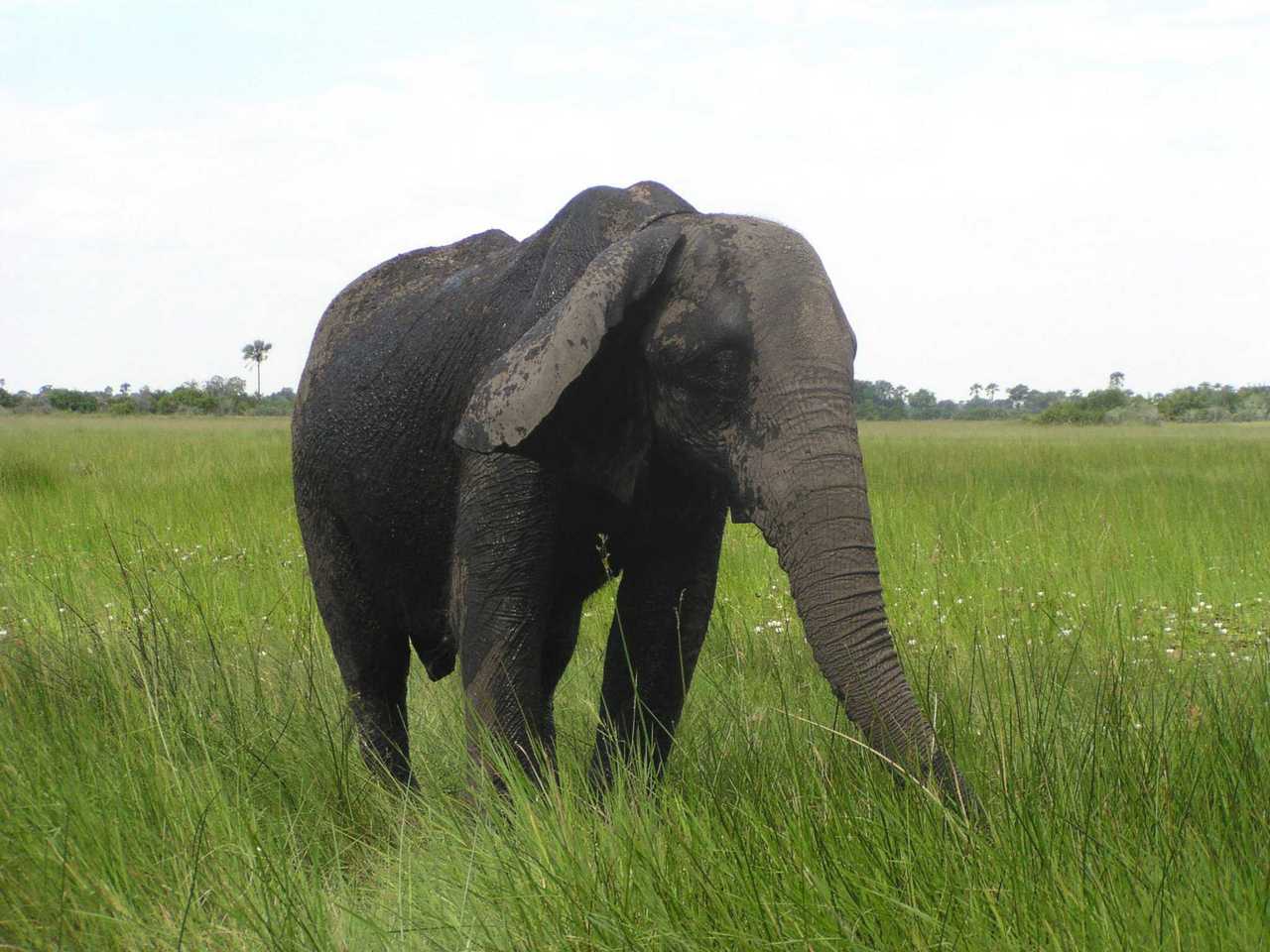}
\caption{Elephant}
\end{subfigure}
\begin{subfigure}[b]{1.4cm}
\includegraphics[width=\linewidth,height=1.4cm]{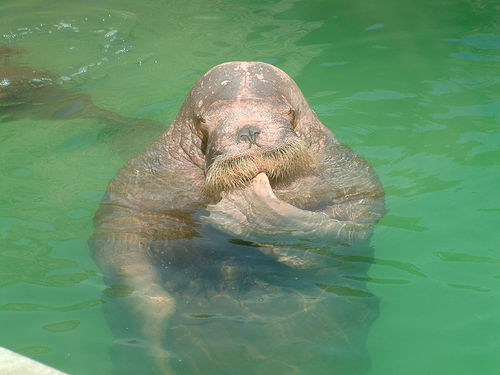}
\caption{Walrus}
\end{subfigure}
\hspace{.3cm}
\rulesep
\hspace{.3cm}
\begin{subfigure}[b]{1.4cm}
\includegraphics[width=\linewidth,height=1.4cm]{figures/blue+whale_0002}
\caption{B. Whale}
\end{subfigure}
\begin{subfigure}[b]{1.4cm}
\includegraphics[width=\linewidth,height=1.4cm]{figures/walrus_0003}
\caption{Walrus}
\end{subfigure}
\begin{subfigure}[b]{1.4cm}
\includegraphics[width=\linewidth,height=1.4cm]{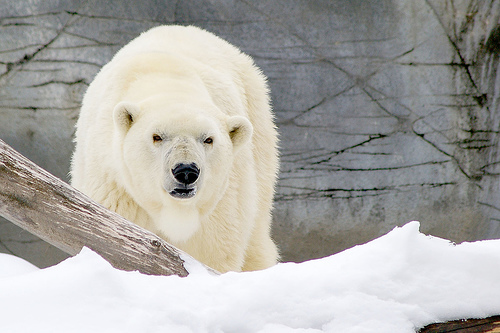}
\caption{P. Bear}
\end{subfigure}
\hspace{.3cm}
\rulesep
\hspace{.3cm}
\begin{subfigure}[b]{1.4cm}
\includegraphics[width=\linewidth,height=1.4cm]{figures/blue+whale_0002}
\caption{B. Whale}
\end{subfigure}
\begin{subfigure}[b]{1.4cm}
\includegraphics[width=\linewidth,height=1.4cm]{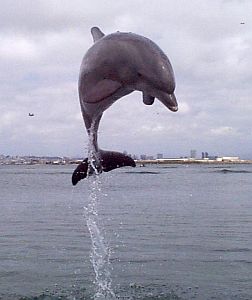}
\caption{Dolphin}
\end{subfigure}
\begin{subfigure}[b]{1.4cm}
\includegraphics[width=\linewidth,height=1.4cm]{figures/walrus_0003}
\caption{Walrus}
\end{subfigure} \\

\begin{subfigure}[b]{1.4cm}
\includegraphics[width=\linewidth,height=1.4cm]{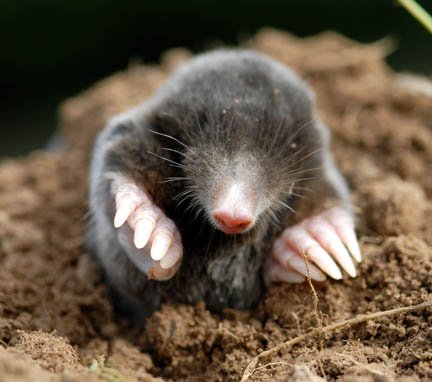}
\caption{Mole}
\end{subfigure}
\hspace{.3cm}
\rulesep
\hspace{.3cm}
\begin{subfigure}[b]{1.4cm}
\includegraphics[width=\linewidth,height=1.4cm]{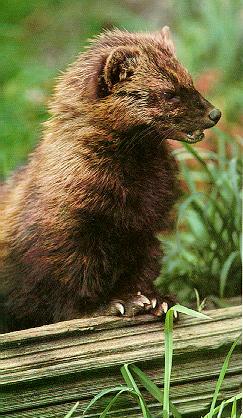}
\caption{Weasel}
\end{subfigure}
\begin{subfigure}[b]{1.4cm}
\includegraphics[width=\linewidth,height=1.4cm]{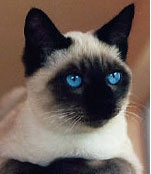}
\caption{S. Cat}
\end{subfigure}
\begin{subfigure}[b]{1.4cm}
\includegraphics[width=\linewidth,height=1.4cm]{figures/blue+whale_0002}
\caption{B. Whale}
\end{subfigure}
\hspace{.3cm}
\rulesep
\hspace{.3cm}
\begin{subfigure}[b]{1.4cm}
\includegraphics[width=\linewidth,height=1.4cm]{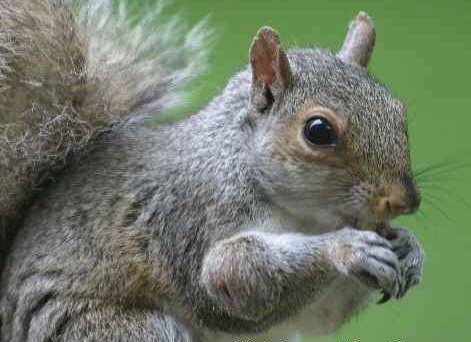}
\caption{Squirrel}
\end{subfigure}
\begin{subfigure}[b]{1.4cm}
\includegraphics[width=\linewidth,height=1.4cm]{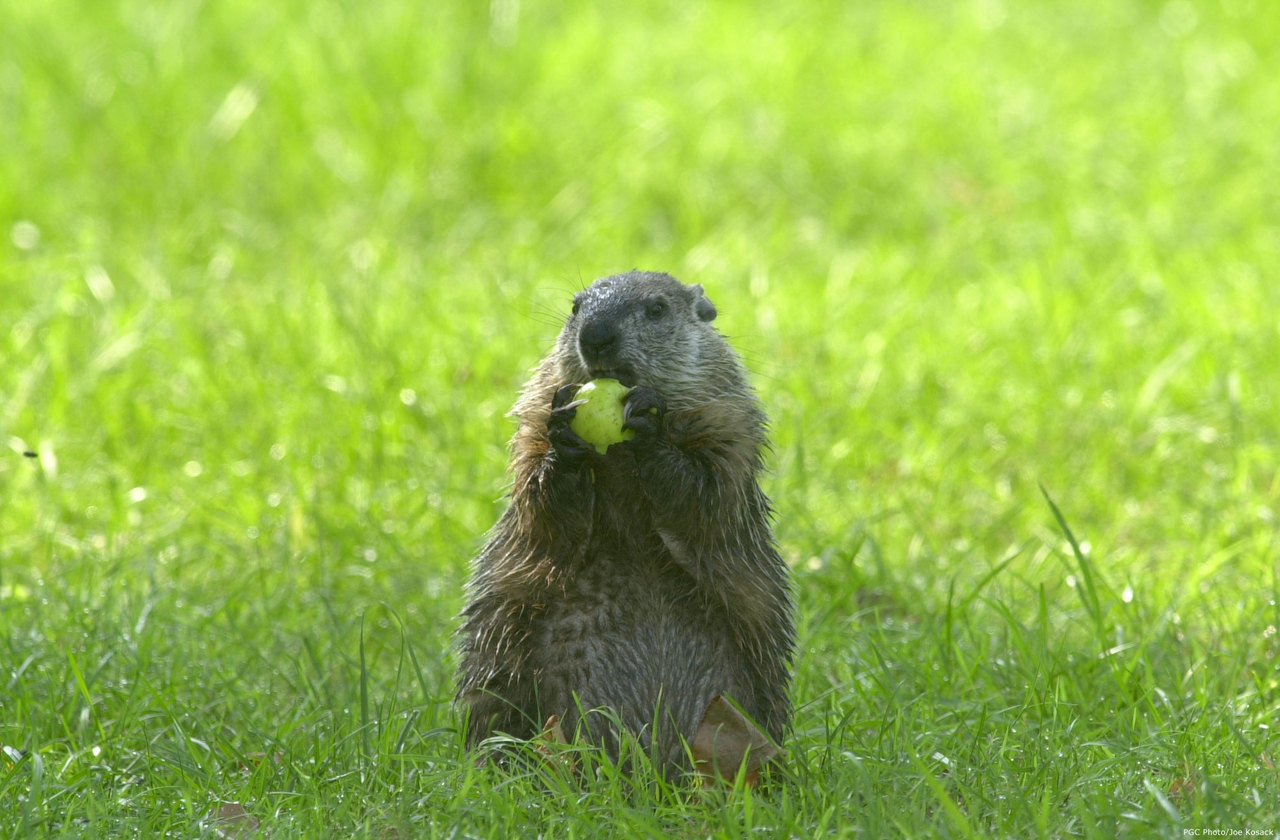}
\caption{Beaver}
\end{subfigure}
\begin{subfigure}[b]{1.4cm}
\includegraphics[width=\linewidth,height=1.4cm]{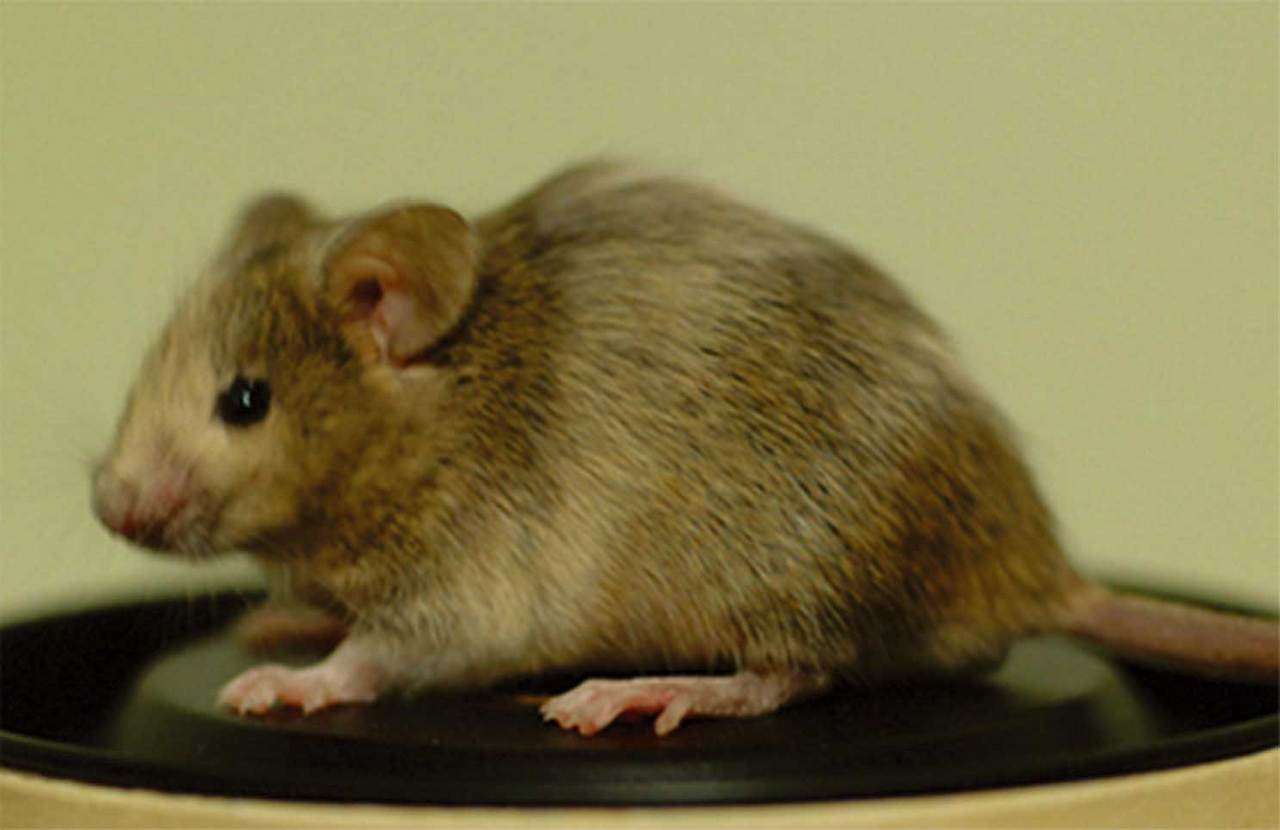}
\caption{Mouse}
\end{subfigure}
\hspace{.3cm}
\rulesep
\hspace{.3cm}
\begin{subfigure}[b]{1.4cm}
\includegraphics[width=\linewidth,height=1.4cm]{figures/mouse_0027}
\caption{Mouse}
\end{subfigure}
\begin{subfigure}[b]{1.4cm}
\includegraphics[width=\linewidth,height=1.4cm]{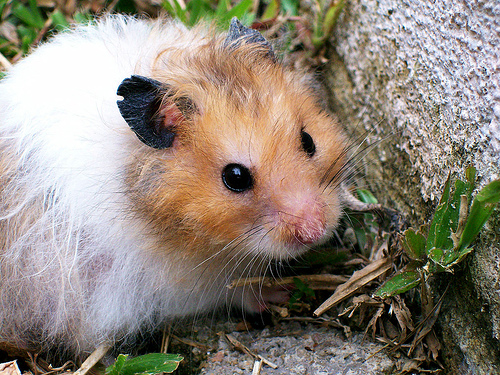}
\caption{Hamster}
\end{subfigure}
\begin{subfigure}[b]{1.4cm}
\includegraphics[width=\linewidth,height=1.4cm]{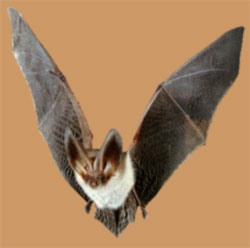}
\caption{Bat}
\end{subfigure}
 \\
\begin{subfigure}[b]{1.4cm}
\includegraphics[width=\linewidth,height=1.4cm]{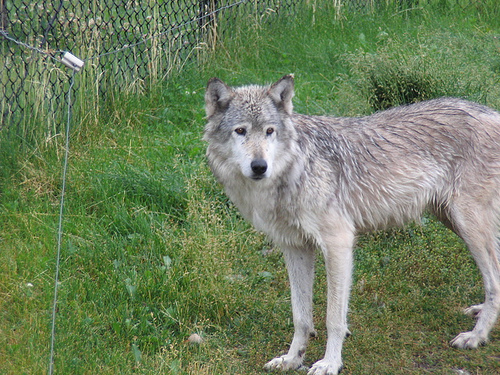}
\caption{Wolf}
\end{subfigure}
\hspace{.3cm}
\rulesep
\hspace{.3cm}
\begin{subfigure}[b]{1.4cm}
\includegraphics[width=\linewidth,height=1.4cm]{figures/polar+bear_0020}
\caption{P. Bear}
\end{subfigure}
\begin{subfigure}[b]{1.4cm}
\includegraphics[width=\linewidth,height=1.4cm]{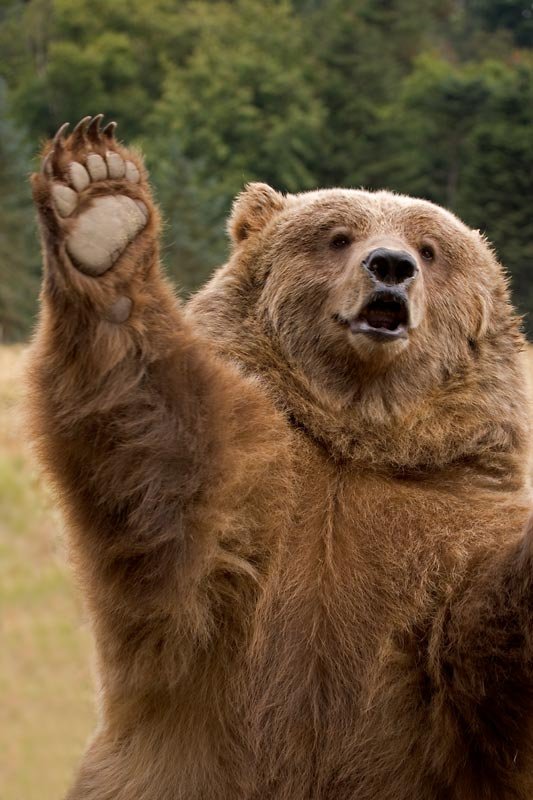}
\caption{G. Bear}
\end{subfigure}
\begin{subfigure}[b]{1.4cm}
\includegraphics[width=\linewidth,height=1.4cm]{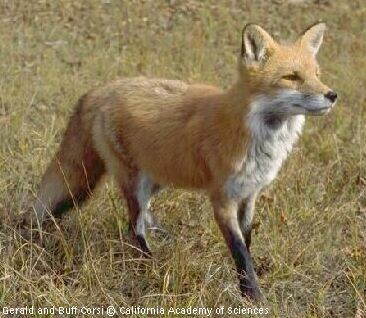}
\caption{Fox}
\end{subfigure}
\hspace{.3cm}
\rulesep
\hspace{.3cm}
\begin{subfigure}[b]{1.4cm}
\includegraphics[width=\linewidth,height=1.4cm]{figures/grizzly+bear_0005}
\caption{G. Bear}
\end{subfigure}
\begin{subfigure}[b]{1.4cm}
\includegraphics[width=\linewidth,height=1.4cm]{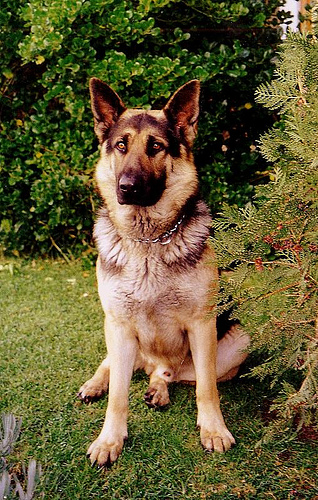}
\caption{Shepherd}
\end{subfigure}
\begin{subfigure}[b]{1.4cm}
\includegraphics[width=\linewidth,height=1.4cm]{figures/fox_0004}
\caption{Fox}
\end{subfigure}
\hspace{.3cm}
\rulesep
\hspace{.3cm}
\begin{subfigure}[b]{1.4cm}
\includegraphics[width=\linewidth,height=1.4cm]{figures/fox_0004}
\caption{Fox}
\end{subfigure}
\begin{subfigure}[b]{1.4cm}
\includegraphics[width=\linewidth,height=1.4cm]{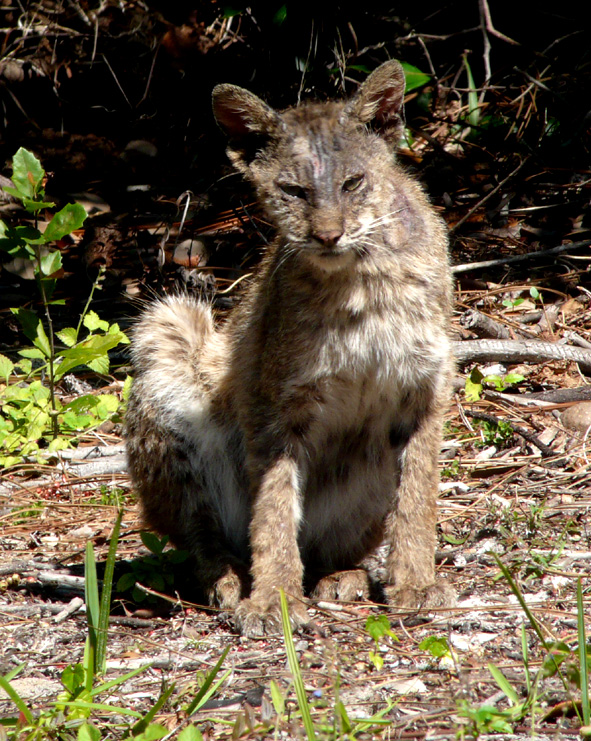}
\caption{Bobcat}
\end{subfigure}
\begin{subfigure}[b]{1.4cm}
\includegraphics[width=\linewidth,height=1.4cm]{figures/german+shepherd_0001}
\caption{Shepherd}
\end{subfigure}

\caption{Top-3 most similar classes for some example classes from the AwA dataset. The similarities of the class word vectors are measured by cosine similarity. The images shown depict class representatives. From left-to-right, the columns show the query class (first column), and the most similar classes according to raw word embeddings (second column), those using the transformation learned by PBT (third column), and those using the transformation learned by IBT (fourth column), respectively.} 
\label{fig:visual_similarity}
\end{figure*}

\vspace{2mm}
\noindent\textbf{Results.} We compare our approach with Objects2Action~\cite{jain2015objects2action} and DAP~\cite{lampert13pami} methods.  The normalized accuracy results are shown in Table \ref{table:comparisonAction}. From these results we see that our approach for relating action names and object cues in the transformed word vector space yields promising results in UCF-Sport dataset. These results show that our embedding transformation function carries substantial semantic information not only between training and test sets, but also across datasets.
\begin{table}
\begin{center}
\caption{Zero-shot action recognition accuracies.} 
\label{table:comparisonAction}

\begin{tabular}{ c c }
\hline
Method & UCF-Sport \\
\hline
    DAP\cite{lampert13pami} &11.7\\
    objects2action\cite{jain2015objects2action} &26.4 \\ 
    Our method&\textbf{28.3} \\
\hline
\end{tabular}
\end{center}
\end{table}

\begin{table}
\begin{center}
\caption{Zero-shot learning using external training class names and their predicate matrices. These \textsc{ext} classes consist of class names outside AwA dataset and do not include image data. The method is trained only on class names and their predicate matrices. We report normalized accuracy.} 
\label{table:outsourced}

\begin{tabular}{lcc}
\hline

Method & Train Classes & Accuracy                    \\
\hline

    \textsc{PBT} & \textsc{ext} & 44.0\\
    \textsc{PBT} & \textsc{AwA} & 60.7 \\
    \textsc{PBT} & \textsc{AwA+ext} & 63.0\\
\hline
\end{tabular}
\vspace{-3mm}
\end{center}
\end{table}

\subsection{Training on Textual Data} As stated before, one of the interesting aspects of our formulation is the ability to train over only textual data (\textit{i.e.} names of attributes, objects and classes), without having any visual examples of training classes.  In this case, using our model, we can use the pre-trained attribute classifiers, together with the learned semantic word vector representation and predict the class of a newly seen example.

To demonstrate the effect, we select 20 classes outside the AwA dataset from Wikipedia Animal List\footnote{\href{https://en.wikipedia.org/wiki/List_of_animal_names}{~en.wikipedia.org/wiki/List\_of\_animal\_names}}, and build an attribute-class predicate matrix. We then learn the corresponding semantic vector space using only these classes that have no image data. The results are shown in Table~\ref{table:outsourced}. Note that, here, we only train the PBT model, because IBT is based on image data.  Training our model using only additional textual class names and their corresponding attribute predicate matrices gives an impressive accuracy of 44.0\%. Moreover, when we augment the AwA train set with these additional class names and their predicate matrix, the accuracy improves from 60.7\% to 63.0\%. These results suggest that the performance of the proposed model can be improved by just enumerating additional class names and their corresponding attribute lists, without necessarily collecting additional image data.

\subsection{Visual Similarities of Word Vectors} One of the favorable aspects of our method is that it can lead to visually more consistent word embeddings of visual entities. To demonstrate this, Figure~\ref{fig:visual_similarity} shows the similarities across the classes according to the original and transformed word embeddings in the AwA dataset.  In the first row, we see that while one of the most similar classes to the \textit{killer whale} is \textit{elephant} using the original embeddings, this changes to the \textit{dolphin} class after using the transformation learned by IBT. We observe similar improvements for other classes, such as \textit{mole} (second row) and \textit{wolf} (third row), 
for which the word embeddings transformed by PBT or IBT training lead to visually more sensible word similarities. 

%
\subsection{Randomly Sampled Vectors} To quantify the importance of initial word embeddings, we evaluate our approach on the AwA dataset by using vectors sampled from a uniform distribution, instead of pre-trained GloVe vectors. In this case, PBT yields 28.6\%, and IBT yields 13.6\% top-1 classification accuracy, which are significantly lower than our actual results (PBT 69.9\% and IBT 60.7\%). This observation highlights the importance of leveraging prior knowledge derived from unsupervised text corpora through pre-trained word embeddings.

%% file: conclusions.tex
\section{Conclusion}
\label{sec:conclusions}
An important limitation of the existing attribute-based methods for zero-shot learning is their dependency on the attribute signatures of the unseen classes. To eliminate this dependency, in this work, we leverage attributes as an intermediate representation, in an unsupervised way for the unseen classes. To this end, we learn a discriminative word representation such that the similarities between class and attribute names follow the visual similarity, and use this learned representation to transfer knowledge from seen to unseen classes. Our proposed zero-shot learning method is easily scalable to work with any unseen class without requiring manually defined attribute-class annotations or any type of auxiliary data. 

Experimental results on several benchmark datasets demonstrate the efficiency of our approach,  establishing the state-of-the-art among the unsupervised zero-shot learning methods. The qualitative results show that the non-linear transformation using the proposed approach improves distributed word vectors in terms of visual semantics. In addition, we show that by adding just text-based class names and their attribute signatures, the training set can be easily extended, which can further boost the performance.